%% file: main.tex
\documentclass[sigconf]{aamas} 

\usepackage{balance} % for balancing columns on the final page

\usepackage{hyperref} % for hyperlinks and colored references
\usepackage{amsmath}
\usepackage{graphicx}
\usepackage{gensymb}
\usepackage{soul}
\usepackage{float}
\usepackage{xcolor}
\usepackage{stfloats}
\usepackage{url}
\usepackage{booktabs}
\usepackage{array}
\usepackage{tabularx}
\usepackage{algorithm}
\usepackage{algorithmic}
\usepackage{newfloat}
\usepackage{listings}

\setcopyright{ifaamas}
\acmConference[AAMAS '26]{Proc.\@ of the 25th International Conference
on Autonomous Agents and Multiagent Systems (AAMAS 2026)}{May 25 -- 29, 2026}
{Paphos, Cyprus}{C.~Amato, L.~Dennis, V.~Mascardi, J.~Thangarajah (eds.)}
\copyrightyear{2026}
\acmYear{2026}
\acmDOI{}
\acmPrice{}
\acmISBN{}

\acmSubmissionID{<<submission id>>}

\title{Towards Real-time Adaptation of Embodied Agent in Human-Robot Collaboration}

\author{Shipeng Liu}
\affiliation{%
  \institution{University of Southern California}
  \department{Department of Electrical and Computer Engineering}
  \city{Los Angeles}
  \state{CA}
  \country{USA}
}
\email{shipengl@usc.edu}

\author{Boshen Zhang}
\affiliation{%
  \institution{University of Southern California}
  \department{Department of Computer Science}
  \city{Los Angeles}
  \state{CA}
  \country{USA}
}
\email{zhehuih@usc.edu}

\author{Zhehui Huang}
\affiliation{%
  \institution{University of Southern California}
  \department{Department of Computer Science}
  \city{Los Angeles}
  \state{CA}
  \country{USA}
}
\email{boshenzh@usc.edu}

\begin{abstract}
Large Language Models (LLMs) have opened transformative possibilities for human-robot collaboration. However, enabling real-time collaboration requires both low latency and robust reasoning, and most LLMs suffer from high latency. To address this gap, we first propose a fine-grained benchmark that explicitly assesses agents' proactive adaptability and temporal responsiveness in the Overcooked-AI environment. Based on evaluation results, we propose MonTA (Monitor-then-Adapt), a hierarchical framework inspired by cognitive science research. MonTA contains three key modules: a lightweight \textit{Monitor} that operates at high frequency (7 Hz) to detect adaptation needs, and two proficient \textit{Adapters} for subtask and path adaptation reasoning that provide instructions to humans at lower frequency. Our results demonstrate that MonTA significantly outperforms baseline agents on our proposed benchmark, achieving superior performance across layouts with varying teaming fluency. User studies confirm the high reasonableness of adaptation plans and consistent language instructions provided by our framework to humans.
\end{abstract}

\keywords{Human-robot collaboration, Large Language Models, Real-time adaptation, Proactive agents, Embodied AI}

\ccsdesc[100]{Human-centered computing~Human computer interaction (HCI)}
\ccsdesc[100]{Computing methodologies~Artificial intelligence}
\ccsdesc[100]{Computing methodologies~Machine learning}
\ccsdesc[100]{Computer systems organization~Robotics}

\definecolor{linkblue}{RGB}{0, 0, 204}

\let\oldcite\cite
\renewcommand{\cite}[1]{\textcolor{linkblue}{\oldcite{#1}}}

\let\oldref\ref
\renewcommand{\ref}[1]{\textcolor{linkblue}{\oldref{#1}}}

\newcommand{\figref}[1]{\textcolor{linkblue}{Figure~\ref{#1}}}
\newcommand{\tabref}[1]{\textcolor{linkblue}{Table~\ref{#1}}}
\newcommand{\secref}[1]{\textcolor{linkblue}{Section~\ref{#1}}}
\newcommand{\appref}[1]{\textcolor{linkblue}{Appendix~\ref{#1}}}
         
\newcommand{\BibTeX}{\rm B\kern-.05em{\sc i\kern-.025em b}\kern-.08em\TeX}

\begin{document}

%%% The following commands remove the headers in your paper. For final 
%%% papers, these will be inserted during the pagination process.

\pagestyle{fancy}
\fancyhead{}

%%% The next command prints the information defined in the preamble.

\maketitle
% Include the separate section files
\input{intro}
\input{related_work}
\input{benchmark}

\input{framework}

\input{results}

\input{conclusion}

%%%%%%%%%%%%%%%%%%%%%%%%%%%%%%%%%%%%%%%%%%%%%%%%%%%%%%%%%%%%%%%%%%%%%%%%

%%% The acknowledgments section is defined using the "acks" environment
%%% (rather than an unnumbered section). The use of this environment 
%%% ensures the proper identification of the section in the article 
%%% metadata as well as the consistent spelling of the heading.

\begin{acks}
% Acknowledgments removed for anonymous submission
\end{acks}

%%%%%%%%%%%%%%%%%%%%%%%%%%%%%%%%%%%%%%%%%%%%%%%%%%%%%%%%%%%%%%%%%%%%%%%%

%%% The next two lines define, first, the bibliography style to be 
%%% applied, and, second, the bibliography file to be used.

\bibliographystyle{ACM-Reference-Format} 
\bibliography{main}

%%%%%%%%%%%%%%%%%%%%%%%%%%%%%%%%%%%%%%%%%%%%%%%%%%%%%%%%%%%%%%%%%%%%%%%%

\input{appendix}

\end{document}

%% file: intro.tex
\section{Introduction}

Robots powered by Large Language Models (LLMs) show great potential for interpreting human instructions~\cite{bubeck2023sparks,ouyang2022training,liu2023llm} and performing tasks through sequential actions in diverse environments~\cite{zhang2023proagent,zhang2024combo,agashe2023evaluating,liu2023llm}. These LLM-powered robots represent a new class of embodied agents~\cite{li2024embodied, zhang2023building}—intelligent systems that interact with their environment through a physical body while leveraging advanced reasoning capabilities for perception, action, and adaptation. The performance of such embodied agents can be assessed by their language understanding, subtask decomposition, and action planning~\cite{li2024embodied}. In highly cooperative, human-involved scenarios~\cite{Liu2023,Liu2024,carroll2020utility}, where agents must interact with humans at every step, e.g. collaborative cooking~\cite{carroll2020utility}, these embodied agents also need real-time adaptation ability to collaborate seamlessly with humans~\cite{weber2011building}. 

Recent LLM-powered agents~\cite{zhang2023proagent,zhang2024combo,agashe2023evaluating,liu2023llm} often operate in a semi-collaborative mode: they await human instructions and execute reactive policies, assuming those instructions and actions are always correct. Consequently, these agents lack proactiveness. Schoenegger et al.~\cite{schoenegger2024ai} demonstrated that state-of-the-art LLMs can often match or exceed human performance across various tasks, suggesting that embodied agents could take the initiative—proactively adapting and guiding humans, particularly those with lower skill levels~\cite{liu2024effect}. However, enabling truly proactive decision-making in embodied agents faces two key challenges: (1) the substantial latency of LLM inference impedes instantaneous reasoning and timely intervention, and (2) the agent must articulate a clear rationale for its instructions and adaptation strategies. Addressing these challenges is crucial to enhancing agent proactiveness and improving performance in dynamic, collaborative settings.

To address these challenges, we focus on enabling embodied agents to make proactive, real-time, and reasonable adaptations and instructions to humans during collaboration. Yet existing evaluation approaches face significant limitations. The recent benchmarks~\cite{agashe2024llmcoordinationevaluatinganalyzingmultiagent, li2024embodied} evaluate the capabilities of embodied agents across several dimensions, including instruction interpretation, subtask decomposition, and action planning, but they are primarily designed for semi-collaborative settings in which humans are assumed to provide explicit instructions, with average response times of 2-5 seconds that are insufficient for real-time collaboration requiring sub-second response times. By contrast, the Overcooked-AI~\cite{carroll2020utility} environment targets real-time human–robot collaboration. Futhermore, Overcooked-AI alone does not sufficiently assess an agent's fine-grained collaborative competencies. Its reliance on game score as the principal evaluation metric restricts the measurement of proactive adaptation and instruction-giving abilities.

To bridge this gap, we design different overcooked environment layouts with different teaming fluency~\cite{hoffman2019evaluating,nikolaidis2024algorithmic} to incorporate different adaptive scenarios. Building on the designed layouts and scenarios, we implement a modular evaluation of an LLM's latency and accuracy across three dimensions: detecting when adaptation is needed, subtask-level adaptation, and action-level adaptation. This evaluation shows a clear speed--accuracy trade-off: relying on rapid responses or smaller models keeps the interaction fast but misses subtle adaptation cues, whereas slower, more deliberate reasoning breaks the sub-second pace needed for smooth collaboration.

Motivated by these findings and drawing on the dual-process theory of human decision making~\cite{kahneman2011thinking}, we mirror how people quickly notice anomalies and pause to think before acting. In the same spirit, we introduce \textbf{MonTA}, a hierarchical framework where lightweight reasoning modules—small or fine-tuned LLMs or embedding-based classifiers—continuously monitor for cues that merit deeper analysis. When such ``stop-to-think’’ moments arise, MonTA escalates to a more deliberate LLM to provide adaptations or targeted guidance. This architecture separates lightweight monitoring (System~1) from high-accuracy adaptation planning (System~2), enabling embodied agents to preserve responsiveness while still providing well-grounded guidance. Our evaluation on the proposed benchmark shows that by integrating systems with complementary capabilities, \textbf{MonTA} achieves comparable or superior performance in proactive adaptation and instruction while maintaining real-time collaboration.

To summarize, our key contributions include:
\begin{itemize}
    \item We developed a modular and fine-grained evaluation framework for assessing agents' real-time proactive adaptation capabilities based on Overcooked-AI~\cite{carroll2020utility}.
    \item We developed \textbf{MonTA}, a hierarchical framework that integrates fast monitoring system 1 and deliberate adaptation system 2 to enable agents to perform proactive real-time adaptations and output natural language instructions in highly human-robot cooperative environments.   
    \item We conducted experiments and user studies that validate MonTA's capabilities to collaborate with humans.
\end{itemize}

The remainder of this paper is organized as follows: Section II reviews related work, Section III presents our benchmark design, Section IV introduces the MonTA framework, Section V presents experimental results, and Section VI concludes with future directions.

%% file: related_work.tex
\section{Related Work}
\subsection{Embodied Agent Benchmark}
Numerous studies have proposed benchmarks for embodied multi-agent systems~\cite{zhang2023building,agashe2023evaluating,chang2024partnr,puig2023nopa}. Several language-based benchmarks, such as \cite{das2018embodied,majumdar2024openeqa}, have focused on question answering, which emphasizes information gathering but does not consider the physical interaction made by embodied agents. Notably, Li et al.~\cite{li2024embodied} introduced the Embodied Agent Interface, a modular framework for evaluating embodied decision-making processes by considering factors beyond overall task performance. However, the designed benchmarks are semi-collaborative scenarios, where humans give a language instruction and agents respond reactively. This limits their ability to evaluate agents' capabilities of proactive adapting and instructing humans during collaboration. The lack of comprehensive benchmarks for real-time proactive adaptation represents a significant gap in the field. The Overcooked environment~\cite{carroll2020utility} is designed for real-time collaboration, but lacks modularity in its evaluation approach. Most overcooked-based benchmarks rely on end-to-end task completion scores, which fail to capture the nuanced decision-making processes required for embodied agents. Accordingly, this work extends Overcooked-AI to evaluate the modularity of embodied agents during real-time human-robot collaboration.
 
\subsection{Real-time Human-AI Collaboration}
Human-AI collaboration has been a long-standing challenge. Prior works study human-AI cooperation in games such as Hanabi \cite{agashe2024llmcoordinationevaluatinganalyzingmultiagent,cui2021k}, diplomacy \cite{doi:10.1126/science.ade9097}, and overcooked \cite{carroll2020utility,fontaine2021importance,strouse2021collaborating}. Several studies leverage LLMs for decision-making tasks in overcooked. For instance, Zhang et al. \cite{zhang2023proagent} use LLMs to infer other agents' intentions and plan subtasks, while Zhang et al. \cite{zhang2024combo} explore long-horizon inference for improved multi-agent cooperation. However, existing approaches fail to address real-time proactive reasoning. Human reasoning operates through a dual-process system: fast, intuitive responses (System 1) for routine decisions and slower, deliberate analysis (System 2) for complex reasoning~\cite{kahneman2011thinking}. In contrast, current LLM-based agents rely on uniform, high-latency reasoning for all decisions, making them unsuitable for real-time collaboration scenarios. Recent work, such as \cite{liu2023llm}, proposed an LLM-based hierarchical framework to follow human high-level instructions, but it still lacks proactive adaptability and the ability to instruct humans during collaboration. Our work addresses this gap by introducing a hierarchical framework that mirrors human dual-process reasoning, enabling real-time proactive adaptation and instruction-giving capabilities.

\begin{figure}[ht]
\centering
\vspace{0.08in}

\centering
\includegraphics[width=1\linewidth]{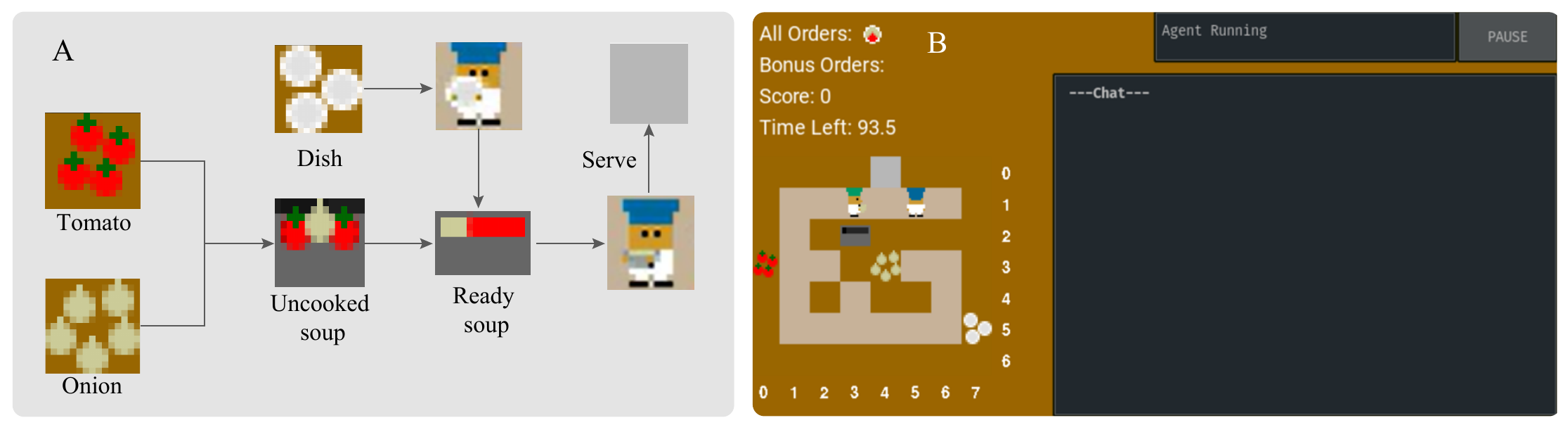}
\caption{The Overcooked-AI simulator (A) The cooking procedure to finish one order. (B) The game interface that we use to test agents and conduct user studies.}
\label{fig:overcook}
\Description{Two-panel figure showing (A) cooking procedure workflow and (B) Overcooked game interface used for testing agents and user studies.}
\end{figure}

\begin{figure*}[ht]
\centering
\vspace{0.08in}

\centering
\includegraphics[width=0.95\linewidth]{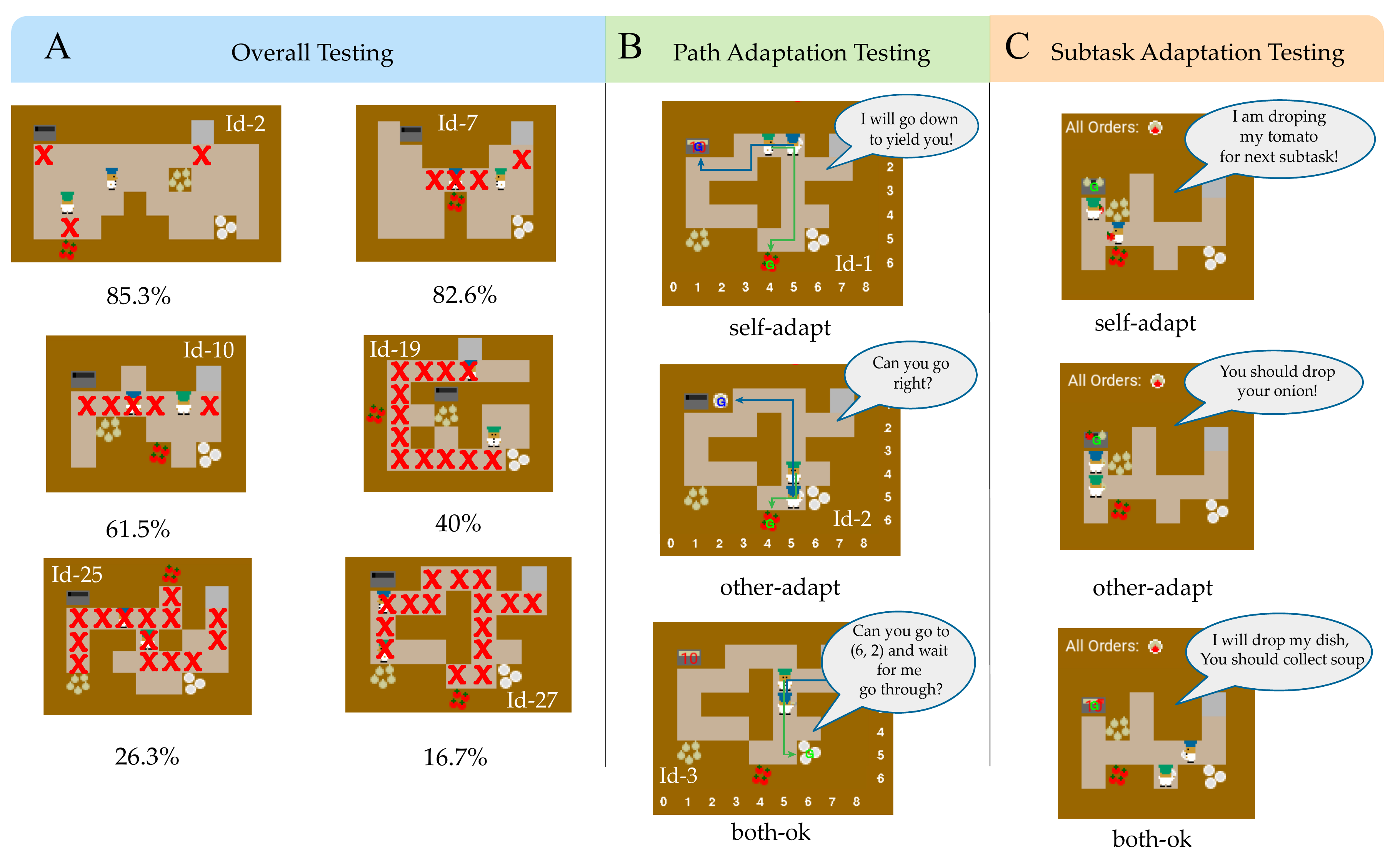}
\caption{
Benchmark for evaluating LLM-based agents' real-time adaptation capabilities. (A) Six selected representative layouts with different teaming fluency from 85.3\% to 16.7\%. The red cross represents a critical point that would interfere with another agent's workflow. (B) Three selected representative path adaptation testing frames designed by human experts: self-adapt, other-adapt, and both-ok types, viewed from the perspective of the blue agent. The subtask goal locations for the blue and green agents are marked as blue "G" and green "G", respectively, with their greedy paths shown as arrowed lines. The blue agent is giving language instruction. (C) Three representative subtask adaptation testing frames where the blue agent is giving language instructions.}
\label{fig:reactive_benchmark}
\Description{Three-panel figure showing (A) six Overcooked layouts with varying teaming fluency, (B) path adaptation testing scenarios, and (C) subtask adaptation testing frames for evaluating LLM-based agents.}
\end{figure*}

%% file: benchmark.tex
\section{An Enhanced Overcooked-AI Benchmark}
To thoroughly evaluate the real-time proactive adaptation and instruction of LLM agents, we extended the original Overcooked benchmark and designed different modular tests. Specifically, we first constructed 22 layouts with varying complexity and coordination demands. Secondly, we implemented a communication panel (\figref{fig:overcook}B) to test the effectiveness of real-time language instruction during human-agent collaboration. Finally, we developed three fine-grained evaluation modules to assess the agent's ability to decide when to make proactive adaptations and how to adapt reasonably. The benchmark consists of three main components: (1) diverse layouts with varying teaming fluency, (2) adaptation scenarios requiring proactive decision-making, and (3) modular evaluation criteria of real-time collaboration. Details are presented in the following sections. 

\subsection{Overcook-AI Environment}
The Overcook-AI environment~\cite{carroll2020utility} is designed to test the coordination skills of multiple agents or human agents. Agents work together in a layout (\figref{fig:overcook}B left) to achieve higher scores by preparing and serving soups within a set time frame, following recipe-specific cooking procedures (\figref{fig:overcook}A) to complete the available recipes. Unlike end-to-end AI agents (e.g., behavior cloning or reinforcement learning), the collaboration process between human and embodied agents can generally be divided into two key steps: determining the current subtask and executing atomic actions to finish each subtask.

The possible subtasks in overcooked environments can be represented by going to (x, y) locations and executing the "interact" action. A general list of subtasks includes: 
\begin{itemize}
    \item \textbf{Preparing Ingredients:} Collect the required ingredients (e.g., onions, tomatoes) and add to the pot.
    \item  \textbf{Cooking:} Cook the ingredients and wait for the timer to indicate the soup is ready.
    \item  \textbf{Serving:} Serve the soup in clean dishes and deliver it to the serving location.
    \item  \textbf{Additional subtasks:} Agents can also decide to drop the items in their hands on any empty counter or pick up items that are placed on any empty counter.
\end{itemize}
 To finish one order, agents have to infer the current state and determine the next subtask based on the recipe. Once the agent determines a subtask and its corresponding position, the agent can be directed through a sequence of atomic actions: \textit{up}, \textit{down}, \textit{left}, \textit{right}, \textit{stay}, and \textit{interact} to finish the subtask.

\subsection{New layouts with different teaming fluency}
In order to better evaluate the agent's real-time adaptation capabilities, we adopted the teaming fluency metrics discussed by Hoffman~\cite{hoffman2019evaluating} and Nikolaidis~\cite{nikolaidis2024algorithmic} to design layouts for our real-time adaptation benchmark. The teaming fluency of a layout is defined as the percentage of non-obstructed areas within the total free area of a layout. If one agent stays at one position and does not move, and another agent cannot finish the task independently, we then mark this position as an obstructed area (red crosses as shown in~\figref{fig:reactive_benchmark}A). With this definition, a higher teaming fluency score suggests an open layout where agents can operate independently without much need to account for each other's presence. Conversely, a lower teaming fluency indicates a more confined and narrow layout, necessitating agents to adapt to one another.

To generate layouts with different teaming fluency, we adopt the following three-step process: (1) we use GPT-4o to generate layouts with symbolic text representation by prompting it to vary the positions of interaction points (e.g., onion, tomato, pot) and adjust the number and positions of empty counters to change the free space. (2) After generating enough layouts, we run a script to filter layouts based on whether they are solvable and teaming fluency. (3) Finally, we manually review the layouts to ensure they are suitable for investigating different adaptation skills and revise them if needed.

We have selected 22 layouts with teaming fluency scores ranging from 88.37\% to 7.14\%. More details are shown in the \appref{app:bench} A.2. These layouts impose constraints on concurrent motions with gradually increasing complexity, requiring agents to adapt to dynamic situations in real-time.

% \subsubsection{Evaluation criteria}For different layouts, we measure the overall score achieved in a certain time threshold as a metric. To achieve good performance in the overall evaluation, the agent needs to show the ability to adapt subtasks as well as low-level paths, especially when paired with non-adaptive agents.

% The overall score reflects the agent's general adaptation capability. However, it cannot independently assess path adaptation and subtask adaptation, providing less insight. To explicitly evaluate the path adaptation capabilities of embodied agents, we carefully design short-horizon scenarios and frames where the subtask of each agent is provided. 

\subsection{Selected Adaptation scenarios}\label{subtask_bech} 
Based on the layouts, we design scenario frames that each of them capture a specific timestep during collaboration between an LLM-agent and a “human agent” (a naive partner lacking adaptive ability). These scenarios assess whether the LLM-agent can adapt its own behavior or proactively instruct the human agent to adjust, thereby evaluating its capacity for both self-adaptation and guidance. In these settings, conflicts may arise either between the agents’ subtasks (\figref{fig:reactive_benchmark}B) or along their paths to task completion (\figref{fig:reactive_benchmark}C). The LLM-agent must reason about whether adaptation is necessary and determine how to act—such as modifying its own subtasks or instructing the human agent to yield.

For each scenario, we begin by choosing layouts where teaming fluency falls below 50\%, then vary each agent's state at this timestep and their current goal subtask to generate scenarios. Specifically, the agent state includes the items held by the agent and the human (e.g., onion, tomato, dish), their initial positions, and the goal subtask is represented by the location of the target counters they aim to interact with.

After generating the scenarios, we run two scripts to check for subtask and path conflicts, respectively. The subtask conflict check is based on the recipes, which can be converted to a Directed Acyclic Graph (DAG) through LLM query~\cite{liu2024effect}. We determine whether adaptation is needed by verifying if the goal subtasks of both agents are valid and whether they are attempting the same subtask. When adaptation is required, the DAG is also used to generate the optimal subtask assignments for the LLM-agent and the human agent given the current state. For the path conflict check, we first compute each agent’s default greedy path using breadth-first search (BFS; \figref{fig:reactive_benchmark}B, green and blue curves) and check for collisions. If a conflict is detected, we apply Conflict-Based Search (CBS)~\cite{10.5555/2900728.2900809} to resolve it by selecting alternative routes for either the LLM-agent or the human, while minimizing path cost.

For both subtask and path conflict cases, we select 41 adaptation scenarios that cover situations where human adaptation is optimal (requiring proactive instruction) as well as those where robot adaptation is optimal. Further details are provided in Appendix~\ref{app:bench}.

\subsection{Evaluation criteria}
% \subsubsection{Score of inferring human intention}
% To evaluate the capability of an LLM to infer human intention, we provide 
\subsubsection{Success rate and latency of identifying the need for proactive adaptation} To assess the LLM-agent's ability to detect when adaptation is needed—whether by the agent itself or by providing instructions to its human partner—we employ a one-shot prompt that produces a structured output containing the complete environment state, current recipe, the current and goal positions of each agent, and each agent's greedy path computed via BFS (Prompt details in \appref{appe:prompt} A1). We then evaluate two metrics: the success rate, $SR_m$, of correctly identifying scenarios with subtask or path conflicts, and the reasoning latency, $L_m$, required to produce those identifications.

\subsubsection{Success rate and latency of proposed adaptation plan}
Evaluating the LLM’s ability to generate correct adaptation plans involves two distinct one-shot prompts—one for subtask adaptation and one for path adaptation (see \appref{appe:prompt} A1). In both cases, the model outputs an alternative plan as a target position $(x,y)$.

For subtask adaptation, we verify whether the subtask associated with the new target position no longer conflicts with other agents’ subtasks defined in the DAG-based recipe. From this, we compute the success rate $SR_{sa}$ and the reasoning latency $L_{sa}$.

For path adaptation, we simulate both the human and the agent following their greedy BFS paths. At each step, the LLM-agent is prompted to determine whether adaptation is needed. If so, it provides an alternative temporary target position for one of the agents to avoid the conflict; once adaptation is deemed no longer needed by the LLM-agent, the agent resumes its greedy path. A scenario is considered successful only if both the human and the agent complete their assigned subtasks within the allotted timesteps. We then measure the overall success rate $SR_{pa}$ and reasoning latency $L_{pa}$ to directly evaluate the model’s path adaptation ability and spatial reasoning.

 \subsection{Trade-off between success rate (SR) and latency }

In real-time human–robot collaboration, an LLM must balance reasoning accuracy with inference speed, as excessive latency can degrade user experience. Prior Human-Robot collaboration studies~\cite{KHASAWNEH2019265} indicate that delays above $\sim$100 ms to 500ms  ($\sim$10 to 2 Hz) hinder fluent coordination, while slower but more accurate reasoning (e.g., 2–5 s) remains acceptable for deliberate planning. This motivates us to quantify the trade-off between accuracy and latency in both subtask and path adaptation scenarios.

\begin{figure}
        \centering
        \includegraphics[width=0.9\linewidth]{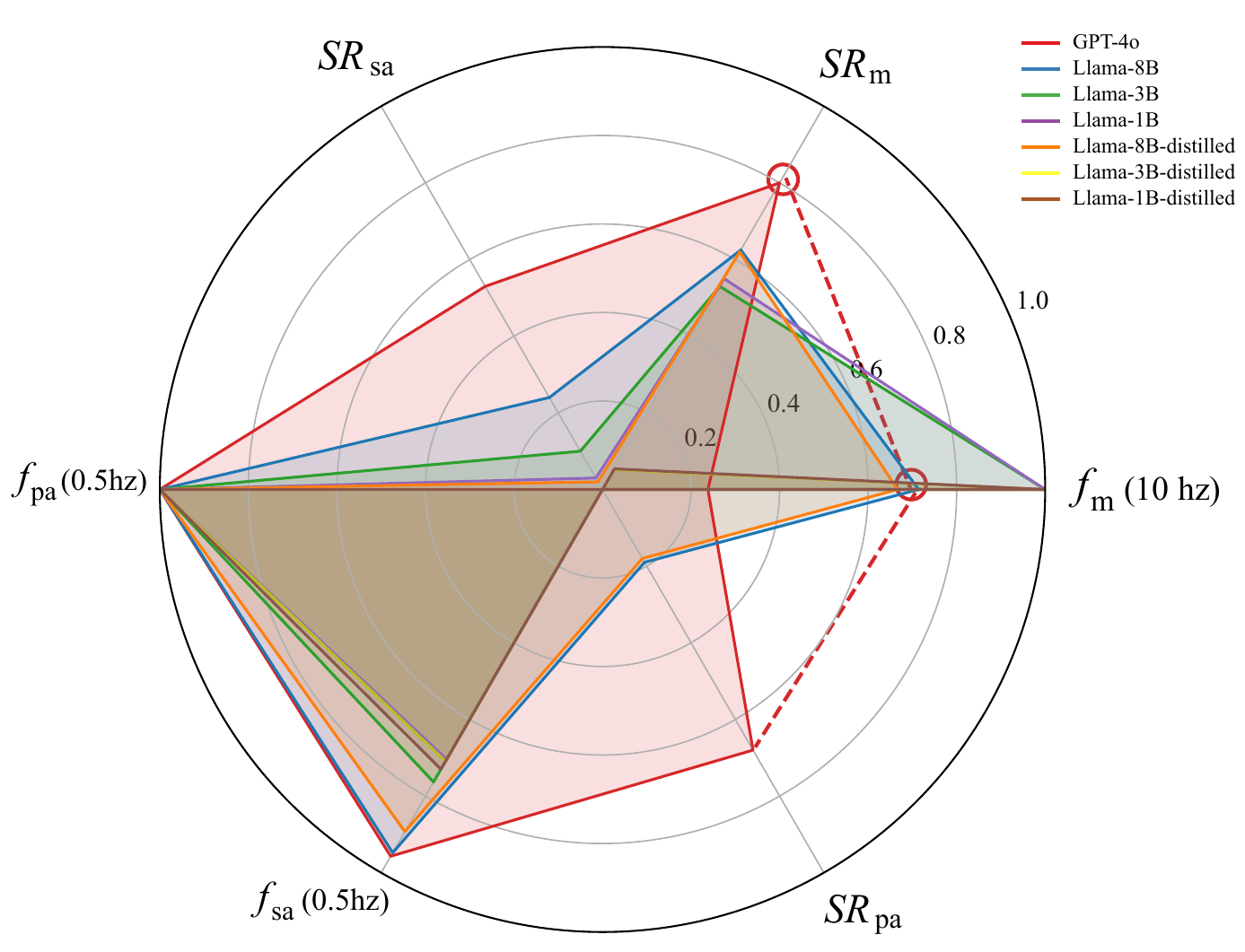}
    \caption{LLM capability and latency evaluation. Success rates are reported for determining whether adaptation is needed ($SR_m$), generating subtask adaptation plans ($SR_{sa}$), and generating path adaptation plans ($SR_{pa}$), along with their corresponding average execution frequencies $f_m$, $f_{sa}$, and $f_{pa}$. Frequencies are normalized to the minimum required for real-time collaboration: 10 Hz for monitoring, and 0.5 Hz for both subtask and path adaptation. Frequencies exceeding these thresholds are cropped. The two circles denote results from the embedding-based classifier, for which adaptation test results are not available.} 
        \label{fig:evaluation}
\Description{Radar chart showing LLM capability and latency evaluation with success rates and execution frequencies for monitoring, subtask adaptation, and path adaptation tasks.}
\end{figure}

To evaluate candidate models, we selected four representatives~\cite{open-llm-leaderboard-v1} differing in size and performance: GPT-4o (via API) and Llama 3.1-8B-Instruct, Llama 3.2-3B-Instruct, and Llama 3.2-1B-Instruct (all running locally through SGLang~\cite{zheng2024sglang} on an RTX 3090). We test the four models in our benchmark scenarios and report both success rate, latency, and processing frequency (the reciprocal of the latency), denoted $f_{m}$, $f_{sa}$, and $f_{pa}$, respectively (\figref{fig:evaluation}).

\begin{table*}[ht]
\centering
\renewcommand{\arraystretch}{1.5} % Adjust line height
\setlength{\tabcolsep}{2pt}

\begin{tabularx}{\textwidth}{l || *{8}{>{\centering\arraybackslash}X}}
\hline
\textbf{Metrics} & \textbf{GPT-4o} & \textbf{Llama-8B} & \textbf{Llama-3B} & \textbf{Llama-1B} & \textbf{Llama-8B-dist.} & \textbf{Llama-3B-dist.} & \textbf{Llama-1B-dist.} & \textbf{text-embedding-3-large} \\
\hline
$L_m$ (s)   & 0.42 & 0.14 & 0.08 & 0.04 & 0.15 & 0.085 & 0.055 & 0.15 \\
$L_{sa}$ (s)& 2.09 & 2.11 & 2.62 & 2.84 & 2.24  & 2.82  & 2.74  & -- \\
$L_{pa}$ (s)& 0.79 & 0.48 & 0.26 & 0.15 & 0.45 & 0.25 & 0.17 & -- \\
$SR_m$ (\%) & 80  & 62.5 & 53  & 55  & 62  & 5.2 & 5.4 & 92 \\
$SR_{sa}$ (\%) & 68  & 19  & 0  & 0  & 18  & 0   & 0   & -- \\
$SR_{pa}$ (\%) & 43  & 24  & 10 & 3  & 2   & 0   & 0   & -- \\
\hline
\end{tabularx}
\end{table*}

Results reveal the expected trade-off: as model size decreases from GPT-4o to Llama 3.2-1B, prompting latency drops substantially (e.g., from 0.42 s to 0.04 s for $L_m$), but success rates decline from 80\% to 52\%. This trend holds across $L_m$, $L_{sa}$, and $L_{pa}$. Moreover, for all models, adaptation plan generation ($L_{sa}, L_{pa}$) is consistently slower than monitoring ($L_m$). Since adaptation plan generation prioritizes reasoning accuracy and can tolerate slower response times (~0.5 Hz) during collaboration, GPT-4o is the only model surpassing the 50\% success threshold for both subtask and path adaptation (where success requires both correct conflict detection and a valid plan generation). Accordingly, we focus on improving the speed of conflict detection while preserving accuracy.

For conflict detection, where speed is critical and higher latency can disrupt collaboration fluency, we explored two approaches. The first aimed to improve smaller models through fine-tuning without sacrificing inference speed. Specifically, we applied Parameter-Efficient Fine-Tuning (PEFT) with LoRA~\cite{hu2021lora} to the Llama-series models. However, this yielded only limited gains, likely because smaller models struggled to distinguish fine-grained adaptation scenarios.

As an alternative, we treated conflict detection as an anomaly detection problem and leveraged GPT-generated embeddings. We encoded each text-represented state using text-embedding-3-large to obtain $O_t$, then computed its maximum cosine similarity with two GPT-4o-labeled embedding sets: one for adaptation scenarios ($A_{t,1:n}$) and one for non-adaptation scenarios ($N_{t,1:n}$). A calibrated threshold determined whether adaptation was required. This embedding-based monitor achieved accuracy comparable to GPT-4o while maintaining an inference frequency of 7 Hz (\figref{fig:evaluation}, red circles), sufficient for real-time operation in Overcooked-AI.

%% file: framework.tex
\section{Real-time hierarchical adaptation framework}
One of the biggest challenges to utilizing LLMs for proactive adaptation and instruction is their high latency, typically ranging from 0.5 to 3 seconds, which is insufficient for real-time collaboration requiring sub-second response times. To enable seamless human-robot collaboration, we must achieve imperceptible latency while maintaining robust reasoning capabilities. 

\begin{figure*}[ht]
\centering
\vspace{0.08in}

\centering
\includegraphics[width=0.95\linewidth]{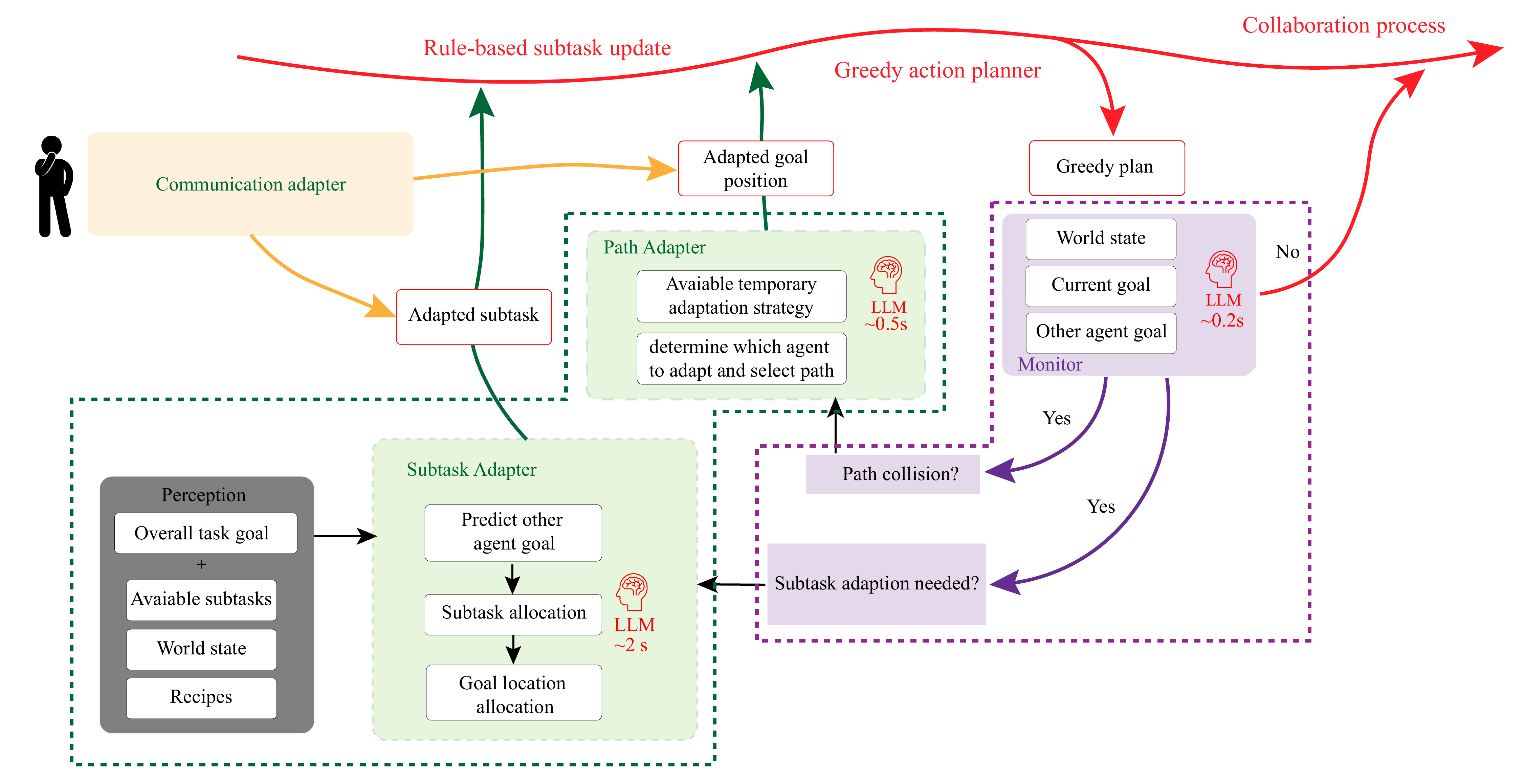}
\caption{MonTA Framework. The framework comprises a real-time monitor and two primary adapter modules: the subtask adapter and the path adapter. The monitor operates at a high frequency to continuously assess the collaboration status and determine whether adaptation is necessary. The adapters are invoked only upon the monitor's request, and they decide how language instructions should be sent to the communication adapter to guide the human collaborator.}
\label{fig:reactive}
\Description{Architecture diagram showing the MonTA framework with a real-time monitor and two adapter modules (subtask and path) for human-robot collaboration.}
\end{figure*}

Inspired by cognitive studies showing that humans interchange between fast and intuitive thinking versus slow and deliberate thinking~\cite{kahneman2011thinking}, we introduce \textbf{MonTA} agent (\figref{fig:reactive}), which leverages a fast and lightweight reasoning module (System 1) to determine if the agent needs to adapt and calls upon a slow but powerful LLM (System 2) to generate detailed adaptation plans. Based on these results, we adopt text-embedding-3-large for System 1 to detect the conflict and GPT-4o for System 2 as the adaptation plan generator. This hierarchical design addresses the latency-accuracy trade-off by separating high-frequency monitoring from low-frequency deliberate reasoning. Specifically, \textbf{MonTA} contains three modules:
\begin{itemize}
    \item \textbf{Monitor}: Operates at high frequency (7 Hz) to continuously assess collaboration status and determine when adaptation is needed and classify adaptation types (subtask vs. path conflicts).
    \item \textbf{Path Adapter}: Invoked by the Monitor when path conflicts are detected. Generates alternative target positions to resolve spatial navigation conflicts and provides language instructions to human collaborators.
    \item \textbf{Subtask Adapter}: Activated by the Monitor when subtask conflicts are identified. Proposes alternative subtask assignments to avoid task allocation conflicts and communicates adaptation strategies to humans.
\end{itemize} 
Details of the framework and each module are shown in the following subsections.
% This section focuses on the design of the framework and how the adapters and monitor enable adaptive collaboration and communication. Detailed prompts for each module can be found in \appref{appe:prompt}A1. 

% MonTA uses a DFS planner to compute path as a sequence of atomic actions, given a target location. These atomic actions include \textit{up}, \textit{down}, \textit{left}, \textit{right}, \textit{interact}, and \textit{stay}. It's worth noting that this planner can be replaced by any existing algorithm, highlighting the generalizability of our proposed framework.
\subsection{Monitor}
% Like System 1 in human cognition, the \textit{Monitor} operates as a fast, intuitive decision-making module that continuously assesses the collaboration environment. It employs lightweight reasoning to determine when adaptation is needed and classifies the type of conflict (subtask vs. path), reserving more deliberate reasoning for the specialized adapters. The Monitor operates at 7 Hz frequency, enabling real-time responsiveness while maintaining computational efficiency.

The Monitor's decision logic follows a hierarchical classification approach: (1) \textit{Subtask-level adaptation} is triggered when agents face conflicting task allocations (e.g., both agents attempting to prepare the same ingredient), and (2) \textit{Path-level adaptation} is activated when spatial navigation conflicts occur (e.g., agents navigating narrow corridors in opposite directions). The Monitor dynamically switches between rule-based greedy planning and adaptive planning based on conflict detection. Once the \textit{Monitor} determines that no further adaptation is necessary, the agent reverts to its original execution plan. This design achieves low inference latency (typically $<$200ms) by using efficient embedding-based classification rather than full LLM reasoning for routine monitoring tasks. Detailed performance evaluation and latency analysis of the Monitor module are presented in \secref{sec:overall_performance}.

\subsection{Subtask Adapter}
The \textit{Subtask Adapter} serves as the System 2 component for deliberate subtask reasoning, activated only when the Monitor detects subtask conflicts. It analyzes the overall task goal, current world state, and recipe requirements to identify alternative target locations and task assignments. Based on evaluation results shown in \figref{fig:evaluation}, we leverage GPT-4o~\cite{openai2024gpt4technicalreport} to implement Chain-of-Thought reasoning~\cite{wei2023chainofthoughtpromptingelicitsreasoning} for complex subtask planning.

The adapter operates in two modes: (1) \textit{Self-adaptation}: The robot modifies its own subtask assignment to avoid conflicts, and (2) \textit{Human instruction}: The robot generates natural language instructions to guide the human collaborator's task allocation. For self-adaptation, the agent uses a greedy planner to execute the alternative target position. When instructing humans, the adapter generates clear, actionable messages such as "Please drop the onion onto the counter at position (x, y)." This dual-mode operation enables flexible conflict resolution while maintaining effective human-robot communication.

\subsection{Path Adapter}
The \textit{Path Adapter} functions as the System 2 component for spatial reasoning, activated when the Monitor detects path conflicts or navigation bottlenecks. Upon receiving a request from the Monitor, it employs GPT-4o's Chain-of-Thought reasoning to evaluate candidate temporary target locations and determine the optimal adaptation strategy. The adapter decides which agent—robot or human—should adapt based on task priority, current positions, and efficiency considerations.

The Path Adapter operates through two execution modes: (1) \textit{Robot adaptation}: The robot uses Breadth-First Search (BFS) to recalculate and execute an alternative route to a chosen temporary position, and (2) \textit{Human instruction}: The robot generates spatial navigation instructions for the human collaborator. For human instruction, the adapter produces clear directional guidance such as "Please yield at position (x,y) so I can pass first" or "Could you take the longer route around the counter?" This approach ensures efficient spatial coordination while maintaining natural communication patterns. An example prompt for path adaptation appears in Appendix~\ref{appe:prompt} A1.

%% file: results.tex
\section{Experiments and Results}

%To further analyze the agents' adaptation behaviors, we tested their success rates in completing specifically designed subtask scenarios using both the greedy planner and our reactive planner with different large language models and investigated the trade-off between adaptation accuracy and decision-making speed. to assess whether humans prefer real-time adaptation and communication capabilities, we conducted human studies involving five participants. Each participant collaborated with MonTA, SAA, and Greedy agents, evaluating their reactive and adaptive capabilities while reporting their preferences. Finally, to further evaluate the effectiveness of the language instructions generated by the MonTA agent, we selected 20 designed frame where adaption is needed and asked the agent to produce adaptive plans. Participants then rated the quality of these generated plans.

\subsection{Performance evaluation}\label{sec:overall_performance}
To demonstrate MonTA's advantages, we compare it against two baselines: a rule-based greedy agent (GA) and a subtask adapter agent (SAA). The GA uses the rule-based subtask planner from~\cite{carroll2020utility} combined with a depth-first search~\cite{tarjan1972depth} and an "auto-unstuck" mechanism that randomly selects from available actions when blocked. The SAA invokes an LLM to propose the next subtask once the current one completes, and then executes each atomic action greedily without any real-time adaptation.

We evaluate MonTA, GA, and SAA on layouts 7, 19, and 27 from our reactive benchmark (\figref{fig:reactive_benchmark}), which have teaming fluencies of 82.6\%, 40\%, and 16.7\%, respectively. For each layout, we pair the target agent (MonTA, GA, or SAA) with a standard GA partner that always follows its greedy plan. We run five trials per pairing and report the average game score with standard deviations. Because the paired GA never adapts or communicates, success hinges solely on the target agent's ability to adapt as layout complexity increases.

\begin{figure}
        \centering
        \includegraphics[width=0.9\linewidth]{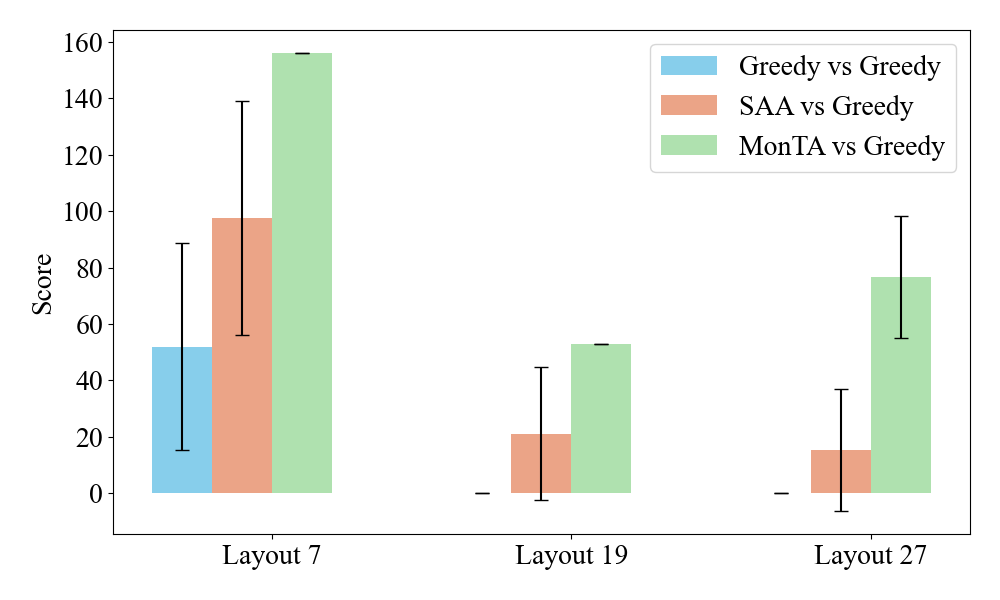}
    \caption{Overall evaluation results. The average score comparison between different agent pairs includes MonTA (ours) v.s. greedy, SAA v.s. greedy, and greedy v.s. greedy. }
        \label{fig:overall}
\Description{Bar chart showing overall evaluation results comparing average scores between MonTA, SAA, and greedy agents across different layouts.}
\end{figure}

\figref{fig:overall} shows that all three agents achieved their best performance on Layout 7 due to its high teaming fluency (82.6\%), which minimizes adaptation needs. As complexity increases and teaming fluency drops to 40\% and 16.7\% in Layouts 19 and 27, GA scores zero across all trials due to ineffective random "unstuck" actions, while SAA suffers performance drops but outperforms GA through LLM-based subtask reasoning. MonTA consistently outperforms both baselines across all layouts, achieving scores of 156 ± 0, 53 ± 0, and 76.6 ± 26.5, with significantly lower variance indicating reliable adaptation detection and execution. This robustness is especially valuable when collaborating with agents of unknown or variable behavior like humans.

\subsection{User study}
True robotic collaborators not only adapt their own behavior but also provide instructions to humans when necessary. In our framework, the agent autonomously adjusts its behavior when self-adaptation is required and sends language instructions only when human needs to adapt.

To evaluate the language instructions and adaptation plans generated by the agent, we recruited 32 volunteers for a human-AI experiment. All volunteers received an introduction to the gameplay mechanics and experiment process, were informed of their rights, and the experiment had departmental approval. Each volunteer was asked to evaluate the reasonableness and consistency of the language instruction generated by the MonTA agent across 20 benchmark scenarios using a 5-point Likert scale. Detailed evaluation criteria are provided in \appref{appe:results} A2.

\begin{figure}
    \centering
    \includegraphics[width=0.9\linewidth]{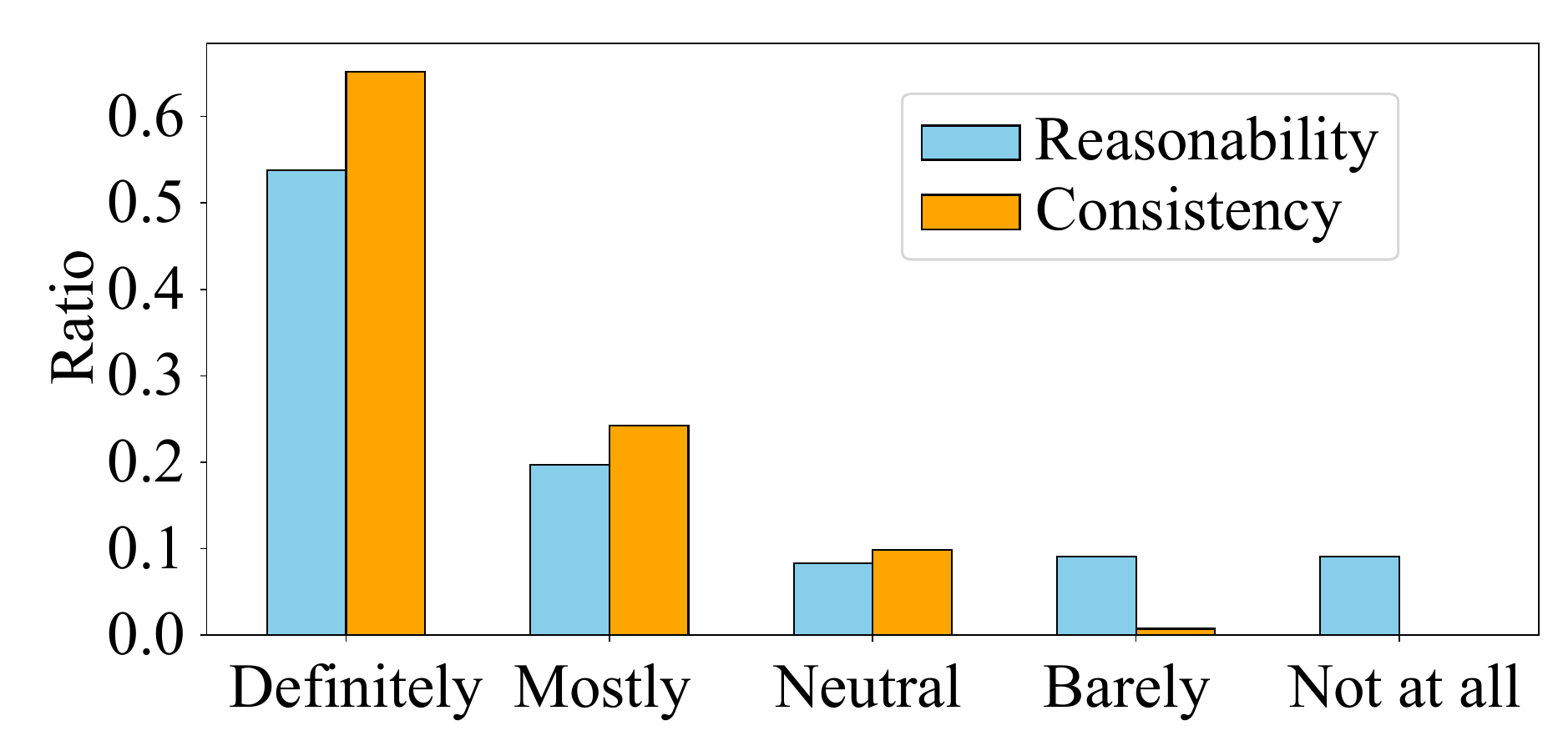}
    \caption{Language instruction evaluation results. Blue and yellow bars show the ratio of LLM instruction reasonability levels and the consistency of LLM suggestions reported by human experts.}
    \label{fig:preference}
\Description{Bar chart showing language instruction evaluation results with blue and yellow bars representing LLM instruction reasonability levels and consistency reported by human experts.}
\end{figure}

The evaluation results, presented in \figref{fig:preference}, reveal that human experts found the suggestions generated by MonTA to be reasonable and consistent in nearly 75\% of scenarios. This indicates that the adapter effectively identifies who should adapt and determines the correct adaptation plan. Such adaptability can reduce the cognitive burden on human collaborators and facilitate seamless human-agent collaboration.

An interesting observation is that in some of the frames, our agent received evaluations with a larger variance. A closer examination of the scenarios revealed divergent human preferences regarding costs and the choice between self-adaptation and other-adaptation solutions. These differences indicate that the optimal solution can vary depending on individual preferences, highlighting the importance of considering human preferences when designing agent instructions. Further details of the evaluation are provided in \tabref{tab:frame_benchmark}.

%% file: conclusion.tex
\section{Conclusion}
% Shipeng original conclusion 
% In this paper, we introduce a benchmark of adaptive scenarios based on the Overcooked-AI environment and MonTA, which uses fast monitoring LLMs and slower, high-accuracy adapters to achieve real-time proactiveness. This allows the monitor to assess the agents' status at a higher rate in real time. When necessary, the LLM transitions to more deliberate "slow thinking" to adapt plans or provide user instructions. Our experiments demonstrate that the real-time adaptation capabilities significantly enhance performance and robustness when two agents collaborate in low-teaming-fluency layouts. Future work should explore adaptive instruction personalization based on user preferences. 

In this paper, we address the critical challenge of enabling real-time proactive adaptation in human-robot collaboration by introducing a comprehensive benchmark and a novel hierarchical framework. Our key contributions include: (1) a modular evaluation framework for assessing agents' real-time proactive adaptation capabilities based on Overcooked-AI, (2) MonTA, a hierarchical framework that integrates fast monitoring using embedding-based classification and deliberate adaptation using GPT-4o, and (3) experimental validation demonstrating MonTA's superior performance and human study results.

MonTA addresses the latency-accuracy trade-off through a dual-process architecture inspired by human cognition: a lightweight Monitor for real-time conflict detection and specialized Adapters for deliberate reasoning when adaptation is needed. Our experiments demonstrate that MonTA consistently outperforms baseline agents across all tested layouts, with significantly lower variance indicating reliable adaptation capabilities. User studies validate the framework's ability to provide natural language instructions, representing a significant advancement toward seamless human-robot collaboration.

Future work should explore adaptive instruction personalization, tailoring guidance to individual user preferences, and extend the framework to more complex multi-agent collaboration. In addition, current foundation models are limited to generating temporary goal positions to avoid conflicts and lack fine-grained understanding of the atomic actions executed by agents. A promising direction is to integrate diffusion models for action-level planning, enabling collaboration strategies to operate at the granularity of individual actions rather than just high-level goals.

% Acknowledgments section removed for anonymous submission

%% file: appendix.tex
\section*{Supplemental Data}

\subsection{A1. Prompt Construction}
We have distinct prompts for Subtask Adapter, Path Adapter, and Monitor. There are two prompts for Monitor as it serves different purposes depending on whether an agent is adapting or following the original greedy path. \label{appe:prompt}

\definecolor{lightlightgray}{gray}{0.9}
\definecolor{codebg}{rgb}{0.95,0.95,0.95}
\lstdefinestyle{customc}{
  belowcaptionskip=1\baselineskip,
  breaklines=true,
  frame=single,
  xleftmargin=\parindent,
  showstringspaces=false,
  basicstyle=\scriptsize,
  keywordstyle=\bfseries\color{black},
  commentstyle=\itshape\color{black},
  identifierstyle=\color{black},
}
\lstset{escapechar=@,
style=customc, 
numbers=none, 
backgroundcolor=\color{codebg},
mathescape=true,
}

\subsubsection{Subtask Adapter prompt}
Subtask Adapter prompt contains environment context, Current game state, filtered actions, and goals. 

\begin{lstlisting}
Context: 
You are a chef that works with another human chef in a kitchen ...
You should follow these rules: ...
The procedure to finish one dish is ...
Recipe book: 
Recipe 0: Requires ingredients: [ingredient 1],[ingredient 2], [ingredient 3] 
=====================
Kitchen state:
    [Kitchen Items in text]
=====================
Your current state:
    1. You are at the coordinates (x,y)
    2. You are facing [item name]
    3. You are holding [item name]
        
The state of the other human chef:
    1. The other chef is at the coordinates (x,y)
    2. They are facing [item name]
    3. They are holding [item name]
=====================
Your available actions: 
    Option 1: [available subtask]
    Option 2: [available subtask]
    ...    
Human available actions:
    Option 1: [available subtask]
    Option 2: [available subtask]
=====================
Goal: 
Your first step will be: analyze the state of the kitchen and items, as well as the recipe to get next best action.select an action from your available actions. and select the target position to interact. choose your target position from kitchen items. do not select target position not listed in kitchen state list.  
Your second step will be: analyze human state,world state and human message/human preference, reason about human intention. choose a human intended position from Delivery location, pot, dispenser listed in kitchen items. do not select target position not listed in kitchen item list.

Return the final data with human intended target position, human intended action id, your final_action_id, target position,
...
\end{lstlisting}

\subsubsection{Path Adapter prompt}
Path Adapter takes information about agents' greedy paths, potential adaptation plans with associated costs, and goals. 

\begin{lstlisting}
Here is your planned greedy path: 
    [agent greedy path]

Human is likely to take following path: 
    [human greedy path]
These two paths overlap path points, which causes collisions.
Your potential adapt plans:
    Plan 1 : [ available plan, plan length]
    Plan 2 : [available plan, plan length]
...
human potential adapts plans:
    Plan 1 : [available plan, plan length]
    Plan 2 : [available plan, plan length]
...

First check the adaptation plan works, by checking if the adaptation plan of one agent will still overlap with the other agent's original path. 
After identifying a valid adaptation plan, choose one with the lowest cost and decide which agent to adapt., please check carefully if the adaptation plan has conflict with other agent's original path
Return the probability of humans adapting with 1 to p_human_adapt if human adaptation has low cost, and the probability of agent adapting with 1 to p_agent_adapt if agent adapt has lowest cost and valid and adapt_index, give me detailed analysis
\end{lstlisting}

\subsubsection{Monitor prompt} 
Monitor has two prompts. One prompt monitoring if agents need to shift to the adaptation path, and one prompt monitoring if agents need to switch back to the original greedy path. We show one prompt here as they only vary in prompt goals. Prompt contains grid layout, agent positions, target positions, agents' greedy path to target, and goals. 

\begin{lstlisting}
Context: 
Grid layout:
This is a 9x7 grid world. The top left corner is (0, 0) and the bottom right corner is (8, 6). Moving down will result second pos coordinates +1, e.g. (0,0) -> (0,1), moving right will results the first pos coordinate +1, e.g (0,0)->(1,0) The Grid contains the following items:
X is obstacle, a is your position, and A is your target position, b is position of human partner and B is human partner's target position(inferred).
[grid layout]

You are at the coordinates: [agent position]
Your target positions:[agent target position]

The other chef is at the coordinates: [others position]
Human Target Position: [others target position]

your planned greedy path:
[agent greedy path]

Human is likely to take following path: 
[human greedy path]
You are current doing a clear temporary adaptation path for collision avoidance:
[adaptation path]

***Your goal: Only using all the information above ***
analyze Do you need to adapt to human behavior? 
for example, you should adapt to human when you want to avoid collision (human current position is on your way). 
otherwise, do not adapt. For example, if you see that both agent are stuck, then it could be good to adapt.

respond your analysis and if you follow greedy or not. respond true if follow greedy, false if not. 
\end{lstlisting}

\subsection{A2. Additional details of benchmark}\label{app:bench}
\subsubsection{Layouts}
The layouts are selected based on the teaming fluency metrics from high to low. Table~\ref{tab:layout_benchmark} provides a visualization of the selected 22 layouts and the corresponding ID as well as the teaming fluency. 
\begin{figure}
    \centering
    \includegraphics[width=0.9\linewidth]{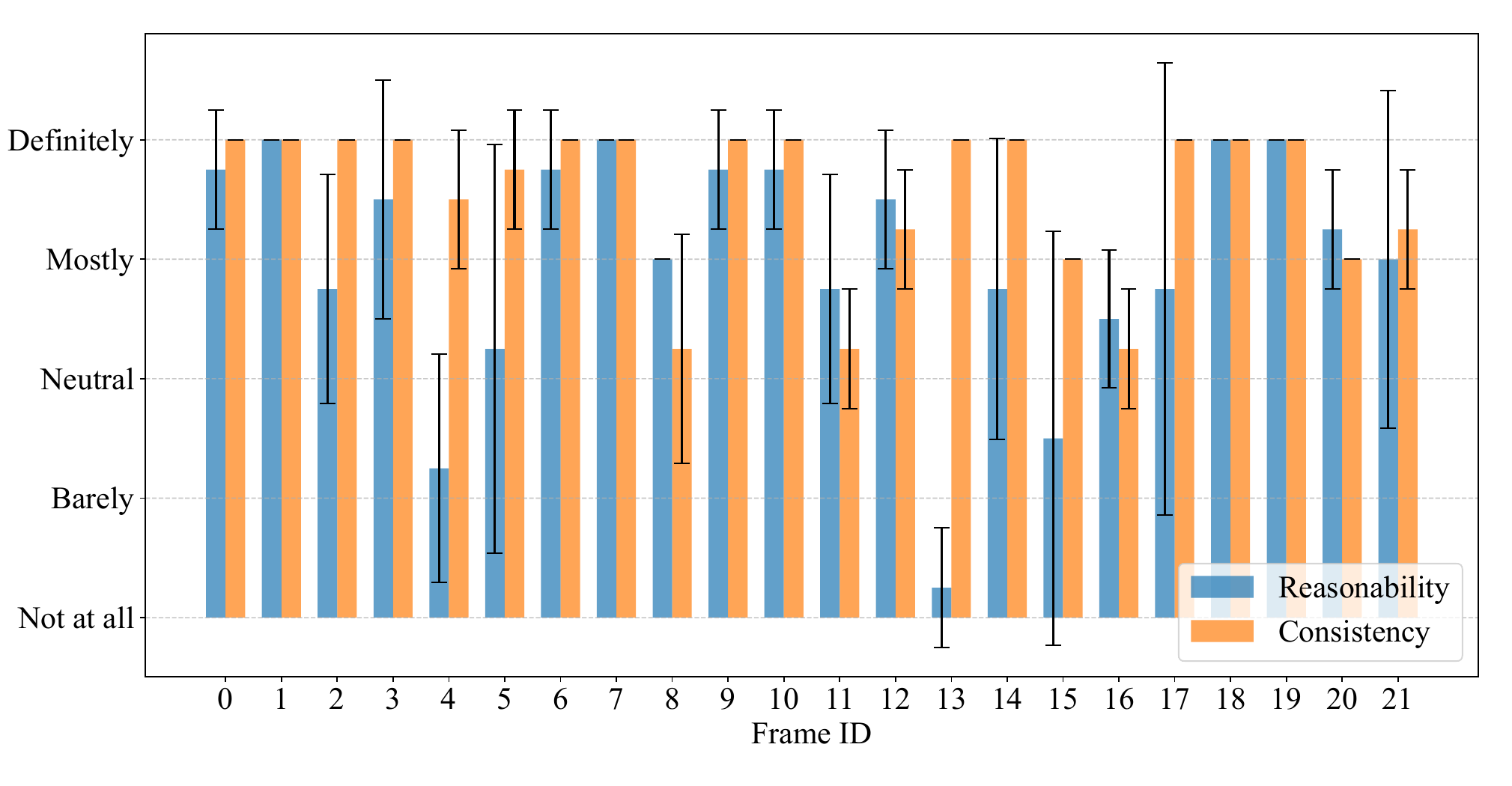}
    \caption{Human reported reasonability and consistency on the llm generated suggestions on each frame. }
    \label{fig:preference_all}
\Description{Detailed chart showing human-reported reasonability and consistency ratings for LLM-generated suggestions across different frames.}
\end{figure}
\subsubsection{Frames}
We generated 41 frames with three types, including self-adapt, other-adapt, and both-ok. We use 20 frames for quantitative testing, which is shown in Table~\ref{tab:subtask_benchmark} and 21 frames for qualitative testing (Table~\ref{tab:frame_benchmark}). 

\begin{table*}[ht]
    \centering
    \caption{All 22 layouts with corresponding number of collision points and team fluency scores.}
    \label{tab:layout_benchmark}
    \begin{tabular}{cccc|cccc}
        \toprule
        \textbf{ID} & \textbf{Layout} & \textbf{Collision Points} & \textbf{Fluency} &
        \textbf{ID} & \textbf{Layout} & \textbf{Collision Points} & \textbf{Fluency} \\
        \midrule
        1 & \includegraphics[width=1.8cm,height=1.8cm]{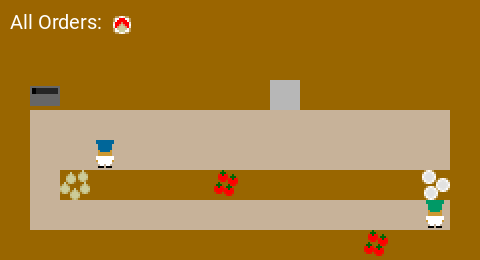} & 5 & 88.37\% &
        2 & \includegraphics[width=1.8cm,height=1.8cm]{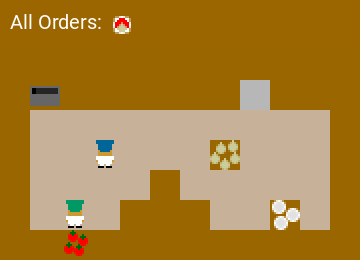} & 5 & 85.29\% \\
        5 & \includegraphics[width=1.8cm,height=1.8cm]{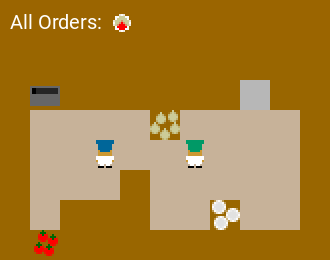} & 6 & 80.00\% &
        6 & \includegraphics[width=1.8cm,height=1.8cm]{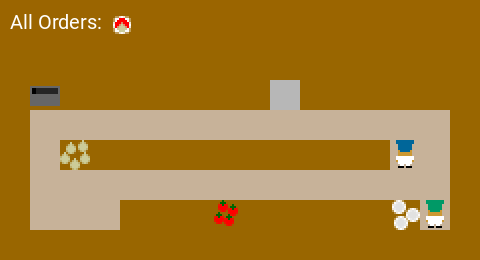} & 5 & 85.71\% \\
        7 & \includegraphics[width=1.8cm,height=1.8cm]{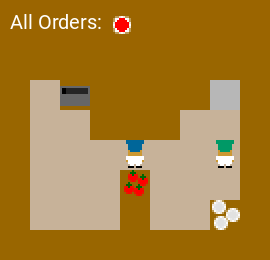} & 4 & 82.61\% &
        8 & \includegraphics[width=1.8cm,height=1.8cm]{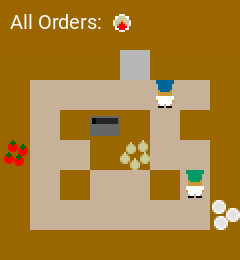} & 5 & 77.27\% \\
        10 & \includegraphics[width=1.8cm,height=1.8cm]{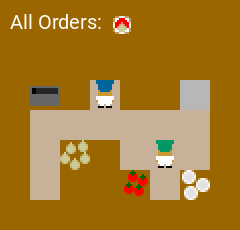} & 5 & 61.5\% &
        11 & \includegraphics[width=1.8cm,height=1.8cm]{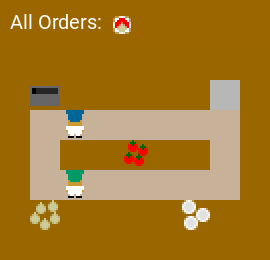} & 5 & 68.75\% \\
        14 & \includegraphics[width=1.8cm,height=1.8cm]{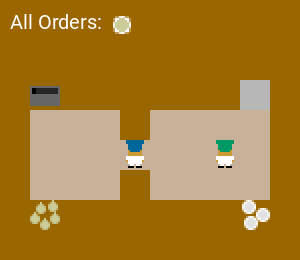} & 7 & 68.18\% &
        15 & \includegraphics[width=1.8cm,height=1.8cm]{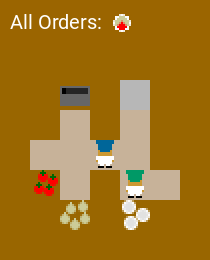} & 7 & 22.22\% \\
        16 & \includegraphics[width=1.8cm,height=1.8cm]{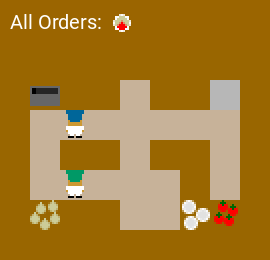} & 5 & 73.68\% &
        17 & \includegraphics[width=1.8cm,height=1.8cm]{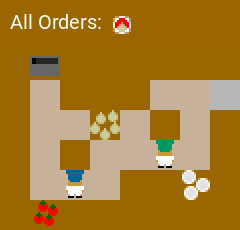} & 9 & 40.00\% \\
        18 & \includegraphics[width=1.8cm,height=1.8cm]{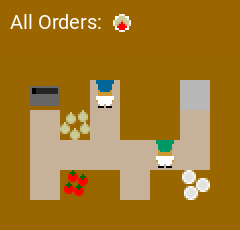} & 7 & 41.67\% &
        19 & \includegraphics[width=1.8cm,height=1.8cm]{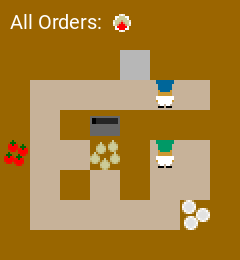} & 12 & 40.00\% \\
        20 & \includegraphics[width=1.8cm,height=1.8cm]{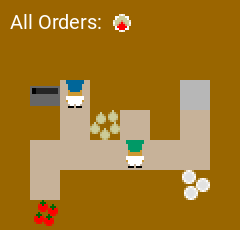} & 9 & 18.18\% &
        21 & \includegraphics[width=1.8cm,height=1.8cm]{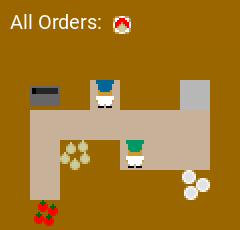} & 8 & 33.33\% \\
        22 & \includegraphics[width=1.8cm,height=1.8cm]{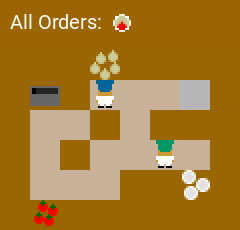} & 13 & 7.14\% &
        23 & \includegraphics[width=1.8cm,height=1.8cm]{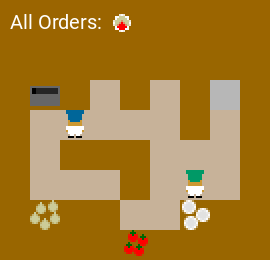} & 12 & 40.00\% \\
        24 & \includegraphics[width=1.8cm,height=1.8cm]{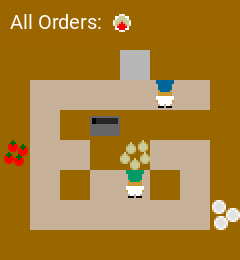} & 12 & 42.86\% &
        25 & \includegraphics[width=1.8cm,height=1.8cm]{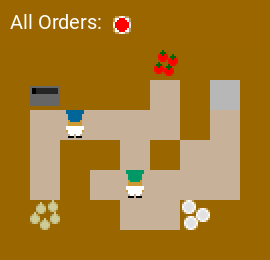} & 14 & 26.32\% \\
        26 & \includegraphics[width=1.8cm,height=1.8cm]{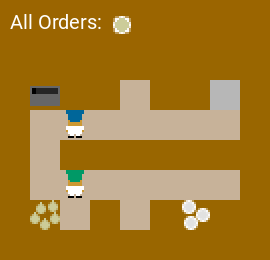} & 15 & 16.67\% &
        27 & \includegraphics[width=1.8cm,height=1.8cm]{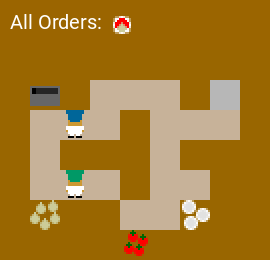} & 15 & 16.67\% \\
        \bottomrule
    \end{tabular}
\end{table*}

\begin{table*}[ht]
    \centering
    \caption{All 21 path adaptation testing evaluation with description and adaptation type.}
    \label{tab:subtask_benchmark}
    \renewcommand{\arraystretch}{0.7}
    \begin{tabular}{c>{\centering\arraybackslash}m{1.6cm}>{\centering\arraybackslash}m{2.5cm}>{\centering\arraybackslash}m{1.3cm}|c>{\centering\arraybackslash}m{1.6cm}>{\centering\arraybackslash}m{2.5cm}>{\centering\arraybackslash}m{1.3cm}}
        \toprule
        \textbf{ID} & \textbf{Frame} & \textbf{Description} & \textbf{Type} &
        \textbf{ID} & \textbf{Frame} & \textbf{Description} & \textbf{Type} \\
        \midrule
        0 & \includegraphics[width=1.6cm,height=1.6cm]{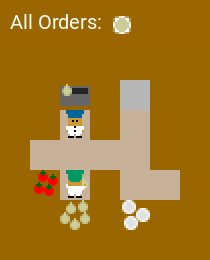} & Blue: pickup onion, Green: pot ingredient & Other-adapt &
        1 & \includegraphics[width=1.6cm,height=1.6cm]{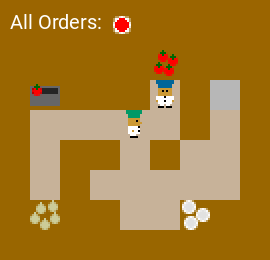} & Blue: pickup onion, Green: pickup tomato & Other-adapt \\
        2 & \includegraphics[width=1.6cm,height=1.6cm]{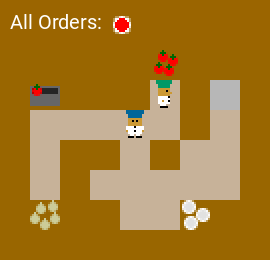} & Blue: pickup tomato, Green: pickup onion & Self-adapt &
        3 & \includegraphics[width=1.6cm,height=1.6cm]{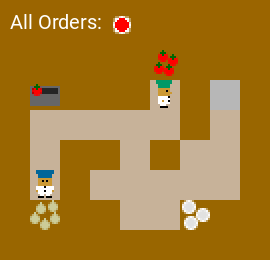} & Blue: pickup dish, Green: pickup onion & Both-ok \\
        4 & \includegraphics[width=1.6cm,height=1.6cm]{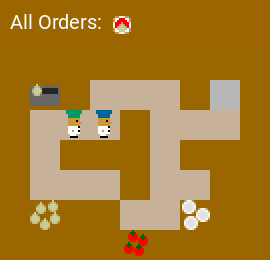} & Blue: pickup onion, Green: pickup dish & Self-adapt &
        5 & \includegraphics[width=1.6cm,height=1.6cm]{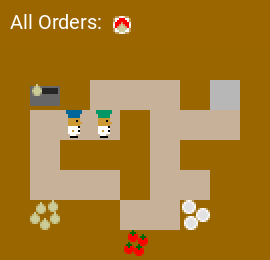} & Blue: pickup dish, Green: pickup onion & Other-adapt \\
        6 & \includegraphics[width=1.6cm,height=1.6cm]{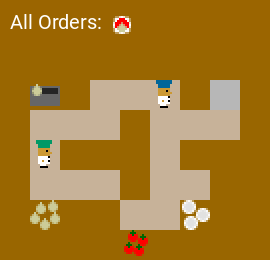} & Blue: pickup onion, Green: pickup tomato & Self-adapt &
        7 & \includegraphics[width=1.6cm,height=1.6cm]{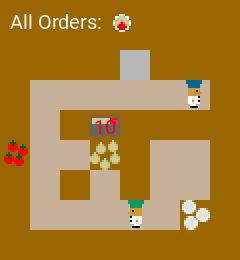} & Blue: pickup tomato, Green: pickup tomato & Both-ok \\
        8 & \includegraphics[width=1.6cm,height=1.6cm]{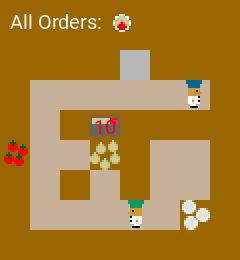} & Blue: pickup onion, Green: pickup tomato & Both-ok &
        9 & \includegraphics[width=1.6cm,height=1.6cm]{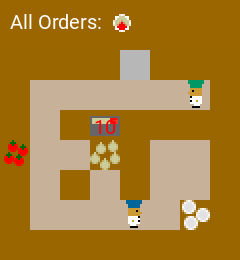} & Blue: pickup tomato, Green: pickup onion & Both-ok \\
        10 & \includegraphics[width=1.6cm,height=1.6cm]{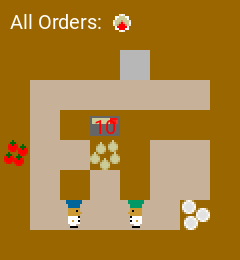} & Blue: pickup dish, Green: pickup tomato & Both-ok &
        11 & \includegraphics[width=1.6cm,height=1.6cm]{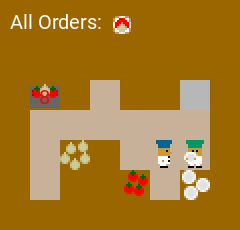} & Blue: pickup onion, Green: pot ingredient & Both-ok \\
        12 & \includegraphics[width=1.6cm,height=1.6cm]{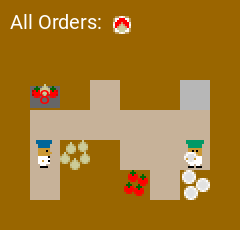} & Blue: pickup tomato, Green: pot ingredient & Both-ok &
        13 & \includegraphics[width=1.6cm,height=1.6cm]{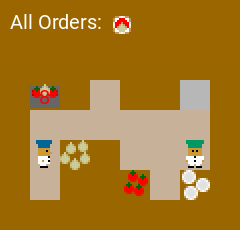} & Blue: pickup tomato, Green: pickup onion & Both-ok \\
        14 & \includegraphics[width=1.6cm,height=1.6cm]{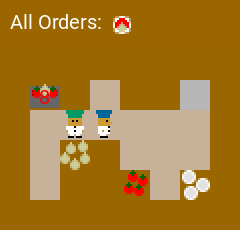} & Blue: pickup onion, Green: pickup tomato & Self-adapt &
        15 & \includegraphics[width=1.6cm,height=1.6cm]{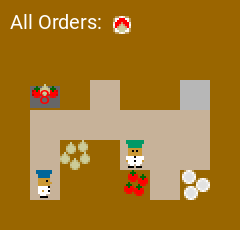} & Blue: pickup dish, Green: pickup onion & Both-ok \\
        16 & \includegraphics[width=1.6cm,height=1.6cm]{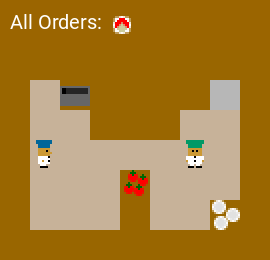} & Blue: pickup dish, Green: pot ingredient & Both-ok &
        17 & \includegraphics[width=1.6cm,height=1.6cm]{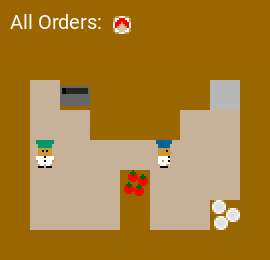} & Blue: pot ingredient, Green: pickup dish & Both-ok \\
        18 & \includegraphics[width=1.6cm,height=1.6cm]{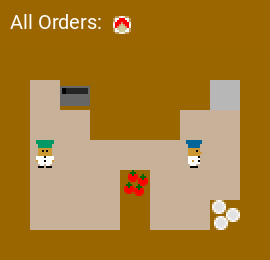} & Blue: pot ingredient, Green: serve soup & Both-ok &
        19 & \includegraphics[width=1.6cm,height=1.6cm]{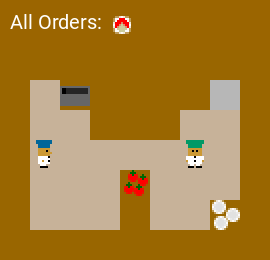} & Blue: serve soup, Green: pot ingredient & Both-ok \\
        20 & \includegraphics[width=1.6cm,height=1.6cm]{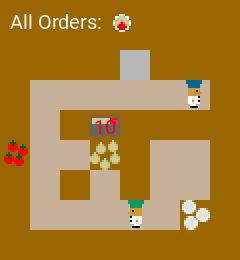} & Blue: pickup dish, Green: pickup tomato & Other-adapt & & & \\
        \bottomrule
    \end{tabular}
\end{table*}

\begin{table*}[ht]
    \centering
    \caption{All 22 frames user study evaluation with descriptions, human rated reasonability and consistency.}
    \label{tab:frame_benchmark}
    \renewcommand{\arraystretch}{0.7}
    \begin{tabular}{c>{\centering\arraybackslash}m{1.6cm}>{\centering\arraybackslash}m{1.8cm}cc|c>{\centering\arraybackslash}m{1.6cm}>{\centering\arraybackslash}m{1.8cm}cc}
        \toprule
        \textbf{ID} & \textbf{Frame} & \textbf{Language Instruction} & \textbf{Reasonability} & \textbf{Consistency} &
        \textbf{ID} & \textbf{Frame} & \textbf{Language Instruction} & \textbf{Reasonability} & \textbf{Consistency} \\
        \midrule
        0 & \includegraphics[width=1.4cm,height=1.4cm]{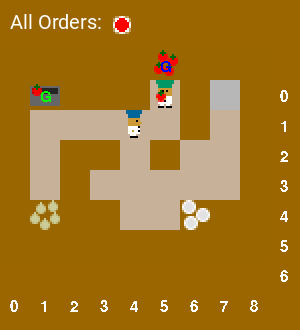} & I will adapt to location [4, 3] & 3.83 (0.41) & 4.00 (0.00) &
        1 & \includegraphics[width=1.4cm,height=1.4cm]{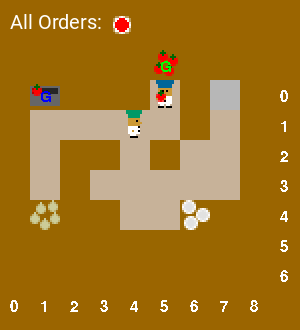} & Could you adapt to location [4, 3]? & 4.00 (0.00) & 3.83 (0.41) \\
        2 & \includegraphics[width=1.4cm,height=1.4cm]{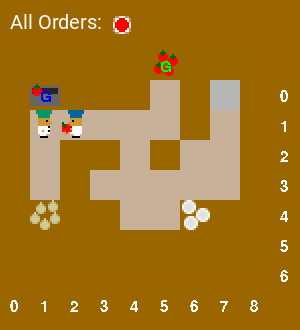} & I will adapt to location [4, 3] & 2.67 (1.21) & 4.00 (0.00) &
        3 & \includegraphics[width=1.4cm,height=1.4cm]{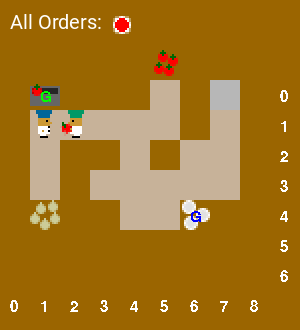} & Could you adapt to location [5, 2]? & 3.17 (1.33) & 4.00 (0.00) \\
        4 & \includegraphics[width=1.4cm,height=1.4cm]{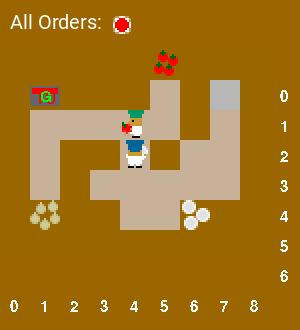} & I will adapt to location [2, 2] & 0.83 (0.98) & 3.50 (0.55) &
        5 & \includegraphics[width=1.4cm,height=1.4cm]{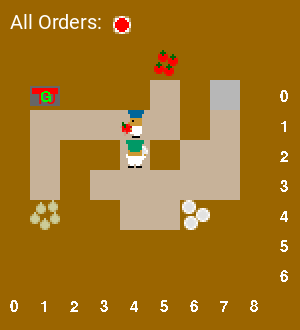} & Could you adapt to location [2, 2]? & 1.67 (1.63) & 3.67 (0.52) \\
        6 & \includegraphics[width=1.4cm,height=1.4cm]{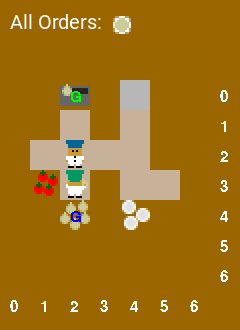} & I will adapt to location [1, 3] & 3.83 (0.41) & 4.00 (0.00) &
        7 & \includegraphics[width=1.4cm,height=1.4cm]{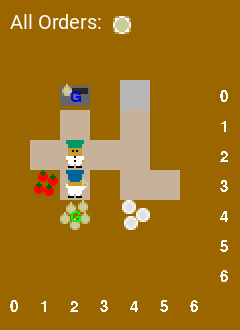} & Could you adapt to location [1, 3]? & 4.00 (0.00) & 3.83 (0.41) \\
        8 & \includegraphics[width=1.4cm,height=1.4cm]{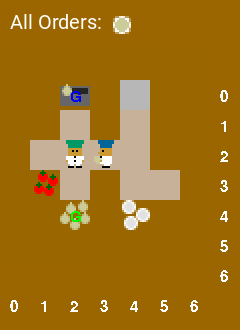} & Could you adapt to location [2, 4]? & 3.17 (0.41) & 2.17 (0.75) &
        9 & \includegraphics[width=1.4cm,height=1.4cm]{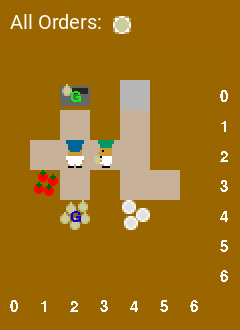} & I will adapt to location [2, 4] & 3.83 (0.41) & 4.00 (0.00) \\
        10 & \includegraphics[width=1.4cm,height=1.4cm]{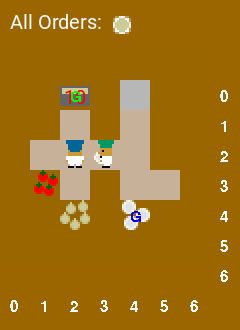} & I will adapt to location [2, 4] & 3.83 (0.41) & 4.00 (0.00) &
        11 & \includegraphics[width=1.4cm,height=1.4cm]{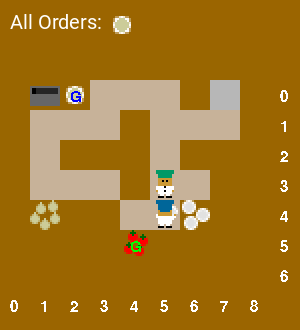} & Could you adapt to location [3, 2]? & 3.00 (0.89) & 2.33 (0.52) \\
        12 & \includegraphics[width=1.4cm,height=1.4cm]{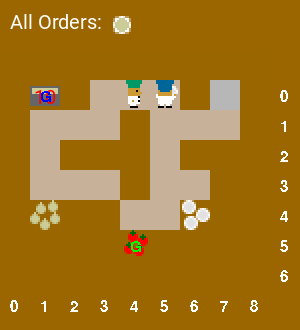} & Could you adapt to location (1, 3)? & 3.50 (0.55) & 3.33 (0.52) &
        13 & \includegraphics[width=1.4cm,height=1.4cm]{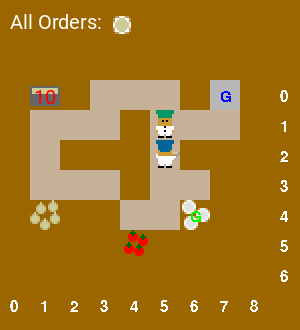} & Could you adapt to [5, 5]? & 0.33 (0.52) & 4.00 (0.00) \\
        14 & \includegraphics[width=1.4cm,height=1.4cm]{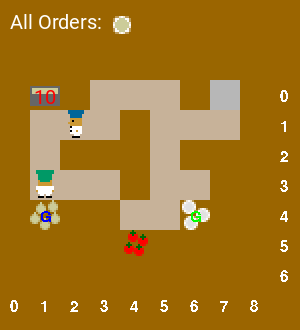} & I will adapt to location [6, 2] & 3.17 (1.17) & 4.00 (0.00) &
        15 & \includegraphics[width=1.4cm,height=1.4cm]{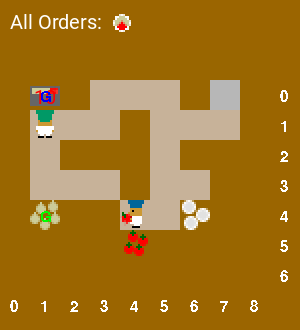} & I will adapt to location (1, 2) & 1.50 (1.64) & 2.67 (0.52) \\
        16 & \includegraphics[width=1.4cm,height=1.4cm]{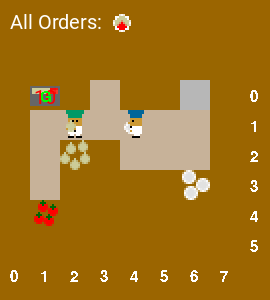} & Could you adapt to location [1, 3]? & 2.50 (1.05) & 2.33 (0.52) &
        17 & \includegraphics[width=1.4cm,height=1.4cm]{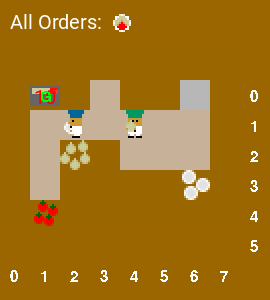} & I will adapt to location [1, 3] & 2.83 (1.60) & 4.00 (0.00) \\
        18 & \includegraphics[width=1.4cm,height=1.4cm]{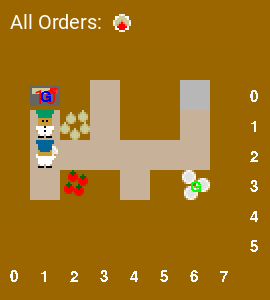} & I will adapt to location [1, 4] & 4.00 (0.00) & 4.00 (0.00) &
        19 & \includegraphics[width=1.4cm,height=1.4cm]{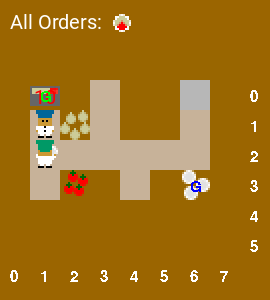} & Could you adapt to location [1, 4]? & 4.00 (0.00) & 3.83 (0.41) \\
        20 & \includegraphics[width=1.4cm,height=1.4cm]{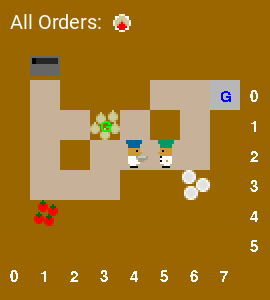} & I will adapt to location [4, 2] & 3.50 (0.55) & 3.00 (0.00) &
        21 & \includegraphics[width=1.4cm,height=1.4cm]{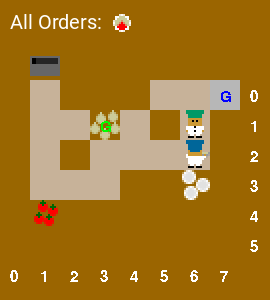} & No adaptation & 2.83 (1.47) & 3.33 (0.52) \\
        \bottomrule
    \end{tabular}
\end{table*}

\subsection{A3. Additional results}\label{appe:results}
Detailed evaluation criteria for the user study are provided here. Participants were asked to rate the reasonableness and consistency of language instructions generated by the MonTA agent on a scale from 1 to 5, where 5 indicates the highest quality.